\documentclass{article}

\usepackage{PRIMEarxiv}

\usepackage[utf8]{inputenc} 
\usepackage[T1]{fontenc}    
\usepackage{hyperref}       
\usepackage{url}            
\usepackage{booktabs}       
\usepackage{amsfonts}       
\usepackage{nicefrac}       
\usepackage{microtype}      
\usepackage{lipsum}
\usepackage{fancyhdr}       
\usepackage{graphicx}       
\graphicspath{{media/}}     

\title{Bench-marking and Improving Arabic Automatic Image Captioning Through The Use of Multi-task Learning Paradigm
}

\author{
  Muhy Eddin Za'ter \\
  University of Colorado Boulder \\
  Boulder, Colorado\\
  \texttt{muhy.zater@colorado.edu} \\
   \And
  Bashar Talafha \\
  University of British Columbia \\
  Vancouver\\
  \texttt{btalafha@student.ubc.ca} \\
}

\begin{document}
\maketitle

\begin{abstract}

The continuous increase in the use of social media and the visual content on the internet have accelerated the research in computer vision field in general and the image captioning task in specific. The process of generating a caption that best describes an image is a useful task for various applications such  as it can be used in image indexing and as a hearing aid for the visually impaired. In recent years, the image captioning task has witnessed remarkable advances regarding both datasets and architectures, and as a result, the captioning quality has reached an astounding performance. However, the majority of these advances especially in datasets are targeted for English, which left other languages such as Arabic lagging behind. Although Arabic language, being spoken by more than 450 million people and being the most growing language on the internet, lacks the fundamental pillars it needs to advance its image captioning research, such as benchmarks or unified datasets. This work is an attempt to expedite the synergy in this task by providing unified datasets and benchmarks, while also exploring methods and techniques that could enhance the performance of Arabic image captioning. The use of multi-task learning is explored, alongside exploring various word representations and different features. The results showed that the use of multi-task learning and pre-trained word embeddings noticeably enhanced the quality of image captioning, however the presented results show that Arabic captioning still lags behind when compared to the English language. The used dataset and code are available at \href{https://github.com/muhi-zatar/CSCI7000_Arabic_Image_captioning}{this link}.  
\end{abstract}


\section{Introduction}
With the enormous surge of social media usage and the continued rapid increase in the visual data generated by users from all around the world, the task of image caption generation is gaining more and more attention and attracting a considerable amount of research efforts from  Natural Language Processing (NLP) community and computer vision community \cite{chen2014learning}. Image captioning; known to be the generation of meaningful captions for images, is a challenging task for machine learning-based models, as they are required to combine both visual and linguistic understanding that includes steps of reasoning to generate high quality captions given an image\cite{elhagry2021thorough}.
However, despite the challenging nature of the image captioning task, recent research endeavors were capable of achieving considerable advances and achievements on this task. Inspired by the recent advances in computer vision and machine translation, researchers have been employing a variety of techniques and algorithms to enhance models performance on image captioning. In addition to that, multiple datasets were developed and collected for image captioning, which helped in accelerating and advancing the research efforts in this field \cite{vinyals2015show, xu2015show, wang2018image, anderson2018bottom, chen2015microsoft}.

However, the overwhelming majority of the available datasets, algorithms, and research efforts were directed toward English, which leaves other languages lagging in this task. Arabic language for instance, is one of the majorly used languages on the web, used by more than 450 million people, and also among internet users, Arabic is the language with the fastest growth rate in the last few years \cite{boudad2018sentiment}. Despite these facts, the image captions generation was untouched by the Arabic vision and NLP communities until recently with only a few shy attempts to create datasets or to apply the advancement in other languages on the Arabic language \cite{al2018deep}.

The Arabic language is known to be rich morphologically, with various distinct dialects that significantly differ from the standard Arabic. These facts impose new challenges for the Arabic NLP, that range from collecting datasets or developing algorithms and models. These challenges are strongly reflected in the task of image captioning, as the attempts so far to create Arabic image captioning datasets are only at best humanly revised translations from the English versions of the datasets such as Flicker30\cite{plummer2015flickr30k}. Also, except for one, these datasets are not available for public, which leads to the absence of reliable benchmarks and hence largely hinges any advancement in this task.

Multi-task learning is not a new idea in computer vision nor NLP, and it has proved its effectiveness in tackling some models' shortcomings and enhancing their performance, also it has been known as a method to overcome the shortage of labeled data for a specific task \cite{zhang2018overview, worsham2020multi}. Also, it has been utilized earlier for better generation of image captions\cite{zhao2018multi, fariha2016automatic}. Therefore, this work attempts to utilize the paradigm of multi-task learning to overcome the scarcity of image captioning data for Arabic, through using models with superior performance on other tasks for different languages as a form of inductive transfer in an attempt to enhance the results of Arabic Image captioning. Also, the datasets developed by translating English captions will be available for public to establish a benchmark for Arabic image captioning to expedite synergy in this task. For the benchmark, this work preliminary results for Arabic image captioning on the well-known translated datasets.

The rest of the document is divided as following. Section 2 lists the related work that tackles Arabic image captioning or uses multi-task learning for image captioning. Section 3 explains the used approach of multi-task learning paradigm and the used datasets in this work. Section 4 illustrates the experimental design in this work, while section 5 presents the results of these experiments. Finally, section 6 concludes the work.

\section{Related Work}
There are few attempts for advancing research in Arabic image captioning through creation of datasets and new algorithms, which will be listed in this section. While also presenting work related to utilizing multi-task learning in image captioning.

The work in the paper \cite{jindal2017deep} in 2017 was the first work to use deep learning for captioning in Arabic language. The approach consists of using deep belief network and root words, in which tree based relations are used to generate captions in the right order. This work was further followed by \cite{jindal2018generating}, in which the author replaces the deep belief networks with recurrent neural networks. However, the work proposed in this document uses the multi-task paradigm to overcome the shortage of data in the Arabic language. 

Furthermore, encoder-decoder paradigm \cite{mualla2018development, al2018automatic, eljundi2020resources} was used for Arabic image captioning, in which it also leveraged pre-trained image models such as VGGNet. This work uses the multi-task learning paradigm which leverages models trained on multiple tasks such as object recognition and detection in addition to other computer vision tasks. While, the work in \cite{eljundi2020resources} publishes the collected data for public use, this work uses different methods for collecting the Arabic captions, as instead of translation using Google API, this work proposes the usage of 3 best evaluated English-to-Arabic open source translation models in order to gain multiple captions per on image.

Work presented in \cite{afyouni2021aracap} suggests a new hybrid approach to tackle the challenging nature of the Arabic language. The hybrid pipeline consists of an object based caption generator, followed by both object detector and attention based caption generator. On the other hand, this work proposes an encoder-decoder architecture, that uses a shared encoder between multiple tasks as an attempt to overcome the shortage of high quality labeled data in Arabic language. 

Multi-task learning paradigm was used for image captioning in \cite{fariha2016automatic, zhao2018multi, wang2018image} for the English language. Also, the work presented in this document differs in the language, the tasks, the proposed architecture from the aforementioned papers and the variety of features that are explored. As the main language in this work is Arabic. Also the architecture proposed in this work is encoder- decoder based architecture with a shared encoder and multiple decoders; one for each task; and the different tasks are trained on a language different than Arabic to tackle the deficiency of data in the Arabic language. 

In summary, the existing literature on Arabic image captioning is novel and only few attempts addressed it, with either applying translation on top of English image captioning systems, or applying famous algorithms on translated datasets or by utilizing some Arabic linguistic features. However, the work lacks a unified dataset that is open for public where each work creates its own dataset, and hence there is an absence of benchmarks for this task.

The contributions are summarized as follows:

\begin{itemize}
    \item An attempt to enhance the performance of Arabic image captioning task by using multi-task learning in order to overcome the scarcity of high quality humanly annotated Arabic dataset for image captioning.
    \item Exploring new different visual and textual features for Arabic image captioning for both multi-task learning and single task learning.
    \item Provide a publicly available translated dataset for Arabic image captioning to compare the different performances of algorithms on this task.
    \item Provide a benchmark for Arabic image captioning in order to accelerate the research in the task.
\end{itemize}

The code and datasets used are available at \href{https://github.com/muhi-zatar/CSCI7000_Arabic_Image_captioning}{this link}.

\section{Method}
This section describes the MTLparadigm, features explored, the architectures and the datasets used to train the image captioning model.

\subsection{Multi-task Learning Paradigm}

The Multi-task Learning (MTL) paradigm can be used for multiple purposes in order to enhance the models' performance on specific tasks. It has various positive effects such as implicit data augmentation, focusing attention on a specific task and can play a regularization role \cite{MTL}.

The concept of MTL consists of hidden layers that are shared between all the trained tasks, and multiple task-specific output layers for each of the tasks. Figure 1 illustrates the MTL paradigm. 

\begin{figure}[h]
    \centering
    \includegraphics[width=3.5in, height=3in]{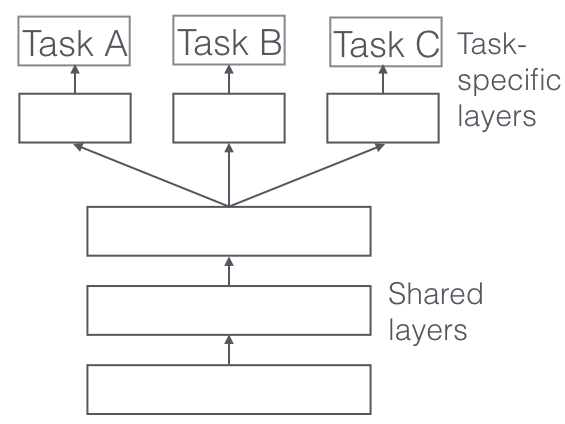}
    \caption{Multi-task Learning Paradigm \cite{MTL}}
    \label{fig:my_label}
\end{figure}

The training methodology of MTL consists of training with each batch each task separately, in which at each batch, the training updates the weights of the task specific layers and the shared layers, while preserving the weights of the other tasks, and with each different batch, the task that is currently being trained changes.

The purpose of using MTL in this work is to overcome the scarcity of high quality data for image captioning in the Arabic Language, therefore multiple language agnostic tasks will be used to enhance both the visual and textual perception of the overall model.

The tasks used will be image captioning, object recognition \cite{deng2009imagenet} and action classification \cite{ge2015action}. The object recognition task is selected in order to assist the model in recognizing the dominant object in an image which in turn will enhance the image captioning model performance in determining the dominant object in an image. Also, the action classification task will improve the image captioning model in recognizing the action in the image which is expected to enhance the image captioning capability of the model.

\subsection{Features}

Different features will be examined for both textual and visual features.

\subsubsection{Textual Features}

Word embeddings are an integral part of most Natural Language Processing tasks, as they act as text encoders in order to provide contextual representation for different words, which usually results in significant enhancements on the performance of the models on different tasks. The following word representations will be used in this work.

\begin{itemize}
    \item \textbf{AraVec} \cite{soliman2017aravec} is a publicly available pre-trained word embeddings for the Arabic language using Word2Vec technique \cite{mikolov2013efficient}. It consists of more than 3.3 Million tokens from different domains such as tweets, Wikipedia and World Wide Web pages.
    \item \textbf{ELMO Based Embeddings} \cite{al2018deep} Elmo embeddings were first to provide different embeddings for the same word depending on the context of the sentence, in order to tackle the issues in Word2Vec models.
    \item \textbf{Bert-Based Embeddings} \cite{alsentzer2019publicly} The Bert-based model uses 12 layers of transformer encoders, in which each layer provides a different representation for the word/sentence. They are trained on huge amount of data, and provide contextual and semantic representations for the words. The pre-trained model in \cite{talafha2020multi} will be used in this work.

\end{itemize}

\subsubsection{Image Features}

The field of computer vision has witnessed significant advances in both the datasets and algorithms, which in turn yielded significant advances in the features that are used as image encoders. In this work, the most used image features will be used for the purpose of evaluating their performance on the task of Arabic image captioning. The following visual features will be used.

\begin{itemize}
    \item \textbf{ResNet} \cite{he2016deep} Residual Networks were first introduced for image recognition and they significantly enhanced the performance of the computer vision models on image recognition task. They have exceptional learning capabilities and therefore it is very common to be used as image encoders and features for different tasks that involve image processing.
    \item \textbf{VGGNet} \cite{simonyan2014very} was first proposed in 2014 as an attempt to investigate the effect of Convolutional Neural Network (CNN) depth effect on the performance of image recognition. It has been widely used as image encoders (features) for different image processing tasks as it has proven its representation capabilities.
\end{itemize}

Both of these networks are trained on ImageNet dataset \cite{deng2009imagenet}.

\subsection{Architectures}

As aforementioned, the MTL paradigm consists of shared layers between all tasks and a task specific layer for each task.

\subsubsection{Shared Layers}
The shared layers that will form the encoder of the MTL paradigm will consist of Convolutional Neural Networks (CNN) layer in order to learn image features based on the actions and objects to be passed for the different decoders.

\subsubsection{Task-Specific Layers}

The task specific layers of the object recognition and action classification tasks will consist of Fully Connected Neural Networks and a Softmax prediction output layer.

Whereas, Long-Short-Term-Memory (LSTM) based decoder will be used as a task specific layer for image captioning due to LSTM capabilities of receiving and generating variable output sequences.

\subsubsection{Action and Object classifiers architecture}

For a more detailed description of the classifiers; each of the object and action classifiers are implemented with a similar architecture. The features will be extracted from either ResNet or VGGNet, that will be fed into CNN Layers, in addition to a Fully Connected Neural Network (FCNN) and finally a FCNN layer with Softmax activation function as an output layer. Figure 2 illustrates the action and object classifiers, whereas table 1 presents the hyper-parameters of the networks.

\begin{figure}[h]
    \centering
    \includegraphics[width=2in, height=4in]{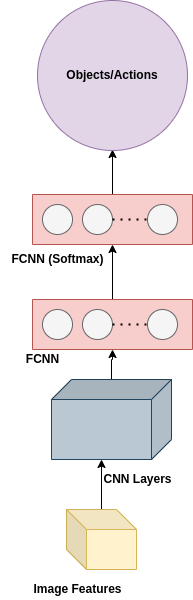}
    \caption{Action and object classifiers architecture}
    \label{fig:my_label}
\end{figure}

\begin{table}[h]
\centering
\caption{Object and Action classifier hyper-parameters}
\begin{tabular}{cc}
\textbf{Hyper-Paramter}                      & \textbf{Value}           \\ \hline
Number of CNN layers                         & 6                        \\
Filter size                                  & 3X3                      \\
Stride                                       & 1                        \\
Number of nodes                              & 64                       \\
FCNN Number of layers                        & 2                        \\
FCNN number of nodes                         & 64                       \\
\multicolumn{1}{l}{FCNN activation function} & \multicolumn{1}{l}{Relu} \\
Dropout                                      & 0.2                      \\ \hline
\end{tabular}
\end{table}

\subsubsection{Image captioning architecture}

The implemented architecture for image captioning is very similar to the architecture in show, attend and tell \cite{xu2015show}. The Encoder-decoder paradigm is utilized, whereas the encoder consists of CNN layers that were fed the image features extracted from ResNet or VGGNet. The decoder on the other hand is implemented using LSTM as they can generate the captions sequentially at every time step. Also Bahdanu \cite{bahdanau2016end} attention was implemented in the decoder which has been proved in multiple domains and tasks that it significantly enhances the performance. Figure 3 illustrates the architecture of the image caption-er, whereas table 2 shows the hyper-parameters used to implement the image captioning architecture.

\begin{figure*}[h]
    \centering
    \includegraphics[width=7in, height=2.5in]{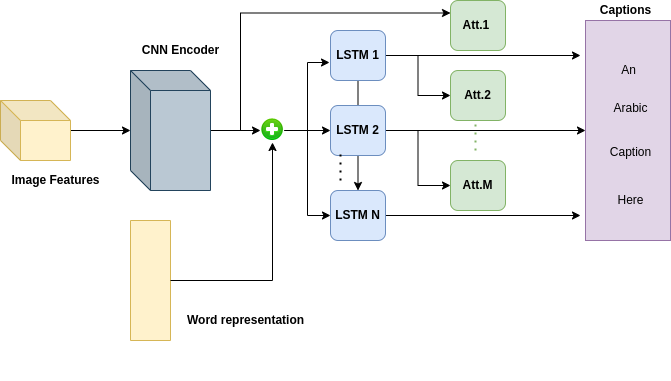}
    \caption{Image captioning architecture}
    \label{fig:my_label}
\end{figure*}

\begin{table}[h]
\centering
\caption{Image captioning architecture hyper-parameters}
\begin{tabular}{cc}
\textbf{Hyper-Paramter}   & \textbf{Value} \\ \hline
Number of CNN layers      & 6              \\
Filter size               & 3X3            \\
Stride                    & 1              \\
Number of nodes           & 64             \\
LSTM Number of layers     & 2              \\
LSTM Number of nodes      & 512            \\
Dropout                   & 0.2            \\
Attention number of nodes & 256            \\ \hline
\end{tabular}
\end{table}

\subsection{Image Captioning Datasets}

The following datasets will be used in the experiments. All the datasets will be translated using 3 open source models and services. These models are Google Translation API, Facebook Machine Translation and University of Helsinki Open translation services.

\begin{itemize}
    \item \textbf{Flickr8K} \cite{flick8k}This dataset consists of 8000 images while each of these images has 5 different captions.
    \item \textbf{Flickr30K}\cite{plummer2015flickr30k} As an extension of the previous dataset, this dataset includes around 32000 images while again each of these images has five different captions. For both of Flickr dataset, the division provided by the authors for training, validation and testing will be followed.
    \item \textbf{MSCOCO} \cite{chen2015microsoft}This dataset has around 120000 for training and validation. While each of these images also has five different captions.
\end{itemize}

The datasets will be combined in both training and testing. However, for speeding the training, 70\% of the training only was used and the same for training. The division is available on the Github repository provided earlier in the document.

\subsection{Other Datasets}

For training the aforementioned action classifiers, Stanford 40 Action \cite{yao2011human} dataset will be used for action classification. The Stanford 40 dataset consists of 9532 different images with 40 classes, in which it has 180-300 images per class.

On the other hand, the object classifier is trained on Cifar-100 dataset \cite{xu2015empirical} which consists of 100 different objects with different levels and depths.

\subsection{Architecture training}

Figure 4 is a summary of the implemented architecture including the three aforementioned tasks and the variation of features described above.

\begin{figure*}[h]
    \centering
    \includegraphics[width=7in, height=5in]{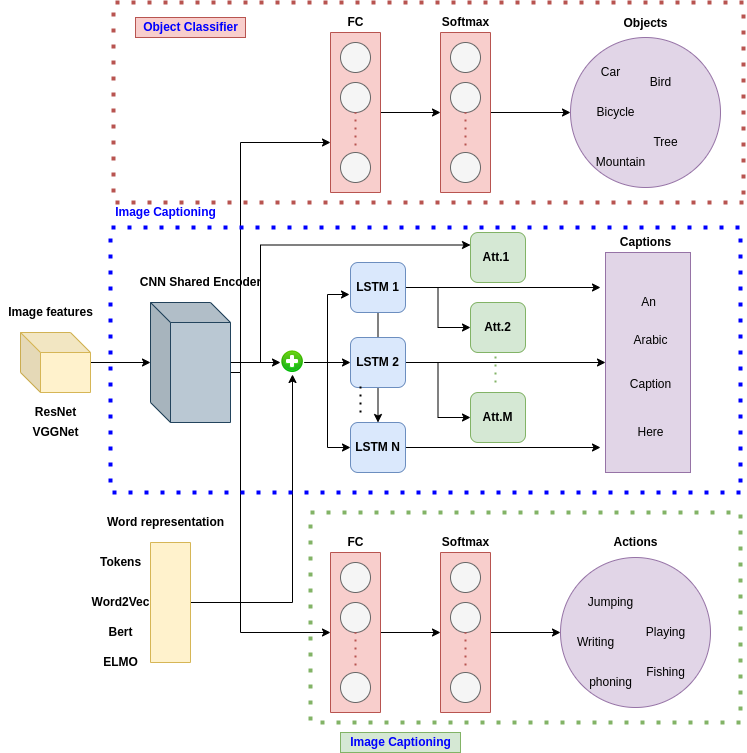}
    \caption{Multi-task learning paradigm for image captioning}
    \label{fig:my_label}
\end{figure*}

Unlike most MTL training designs, which include switching between the tasks after each batch, this work trains the action classifier, then the object classifier, and uses the resulting encoder for image captioning. this training procedure was followed because this work only focuses on enhancing the image captioning performance.

The training related parameters used were Adam \cite{kingma2014adam} as an optimizer, 0.001 as learning rate, batch size of 32 and 80 epochs were trained.

\section{Experimental Design}
This section presents the evaluation metrics and the different experiments conducted to study the effect of each feature and technique on the overall performance.

\subsection{Evaluation metrics}

The following metrics will be used to assess the performance of the different models:

\begin{itemize}
    \item \textbf{BLEU-N} \cite{papineni2002bleu} This is a common metric for evaluating translation models due to the fact that it correlates with human judgment. While using BLEU, the recall and precision are calculated by best match length and n-gram precision.
    \item \textbf{METEOR} \cite{banerjee2005meteor} This metric was also first proposed for the evaluation of translation. It was proposed to tackle the issues of the BLEU score. This metric uses the balanced mean of the unigram precision and recall and the penalty function for incorrect word order.
\end{itemize}

\subsection{Baseline model}

For the purpose of evaluating the effect of each component on the overall performance of image captioning in the Arabic language; at first a baseline model which consists of ResNet as feature extractor for images, a simple tokenizer as word representation, and a CNN encoder solely trained on image captioning without other tasks.

\subsection{Effect of Different Features}

Both the effect of visual and textual features will be evaluated. Different variations of the textual and visual features will be evaluated.

\subsubsection{Text Features}

As they contributed to increased performance enhancement on each task they were used in, text-based features namely word embeddings or representations will be used as text encoders. Aravec, BERT-based embeddings and ELMO-based embeddings.

\subsubsection{Image Features}

The popular ResNet and VGGNet are used as image features in this work. They are compared in order to study the effect of using different image features on image captioning.

\subsection{Effect of Multi-task Learning}

The effect of the addition of each task on image captioning will be evaluated. At first, the shared encoder will be trained on action classifier then image captioning to evaluate the effect of adding the action classifier task. The same pipeline will be repeated for object detection task. Finally, to assess the effect of both tasks; the CNN shared encoder will be trained on action classification, then object classification and eventually image captioning.

\subsection{Comparison with other work and languages}

To evaluate the overall status of the Arabic Image captioning, it is compared against the results on the same datasets in the English language. However, as mentioned earlier in the document, Arabic image captioning results on a unified dataset are not available.

\section{Experimental Design and Results}
Presenting the experimental results of each of the experiments described above. Also, the features and parameters of the baseline model are used when studying the effect of one feature.

Table 3 presents the results of the baseline model described in the above section.

\begin{table}[htb]
\centering
\caption{Baseline model Performance}
\begin{tabular}{ccccc}
\textbf{Feature} & \textbf{B-2} & \multicolumn{1}{c}{\textbf{B-3}} & \multicolumn{1}{c}{\textbf{B-4}}  & \multicolumn{1}{c}{\textbf{METEOR}} \\ \hline
Baseline model           &       34.13          &    23.5             &                             14.31        &         13.13                                                         \\ \hline                                    
\end{tabular}
\end{table}

Table 4 illustrates the results of using different set of work representations.

\begin{table}[htb]
\centering
\caption{Word embeddings effects on the performance}
\begin{tabular}{ccccc}
\textbf{Feature}  & \textbf{B-2} & \multicolumn{1}{c}{\textbf{B-3}} & \multicolumn{1}{c}{\textbf{B-4}} & \multicolumn{1}{c}{\textbf{METEOR}} \\ \hline
Tokenizer           &       34.21          &       23.5          &                               14.31      &   13.13                                                                      \\
AraVec           &     \textbf{35.22}            &        \textbf{24.92}         &                            \textbf{ 14.87}        &           \textbf{13.79}                                                               \\
BERT Based       &        34.46         &    24.15             &                            14.45         &           13.36                                                               \\
ELMO             &      34.33           &    23.98             &                    14.35                 &           13.21                           \\ \hline                                    
\end{tabular}
\end{table}

It is evident that the use of word embeddings has a positive effect on the performance. However, Aravec based embeddings achieved the results as they are trained on the largest amount of data between the three presented techniques. Also, the used ELMO and BERT embeddings were trained on the dialectic version of the Arabic language, which might have negatively affected the results.

Similarly, table 5 shows the results for using different image extraction features. Tokenizer as word embeddings were used in the following table.

\begin{table}[htb]
\centering
\caption{Image features effects on the performance}
\begin{tabular}{ccccc}
\textbf{Feature} & \textbf{B-2} & \multicolumn{1}{c}{\textbf{B-3}} & \multicolumn{1}{c}{\textbf{B-4}}  & \multicolumn{1}{c}{\textbf{METEOR}} \\ \hline
ResNet           &       34.21          &        \textbf{23.5}         &   \textbf{14.31}                                  &   \textbf{13.13}                                                                       \\
VGGNet           &        \textbf{34.22}         &     23.4            &  14.26                                    &     13.12                                                                     \\ \hline                                    
\end{tabular}
\end{table}

There was not any evident difference between the use of ResNet or VGGNet on the results.

Finally, the effect of multi-task learning is presented in table 6. The word embeddings used in this experiment are Aravec as they provided the best results. ResNet is used as the image feature extractor.

\begin{table}[htb]
\centering
\caption{The effect of different tasks on the performance}
\begin{tabular}{ccccc}
\textbf{Task}              & \textbf{B-2} & \textbf{B-3} & \textbf{B-4} & \textbf{METEOR}\\ \hline
Object Recognition    &     35.29                 &      25.23           &             14.91    &         14                         \\
Action Classification &     35.59                 &      25.83           &       15.02          &         14.21                         \\
Both Tasks             &    \textbf{36.38}                  &    \textbf{25.93}             &         \textbf{15.09}        &  \textbf{14.23}                               \\
Single-task Learning  & \multicolumn{1}{l}{34.13} &  23.5               & 14.31                &                     13.13             \\ \hline
\end{tabular}
\end{table}

Multi-task learning achieved better results than single task learning. However, the action classification task achieved better results than object classifier, which can be attributed to the fact that actions are more relevant to image captioning, and the fact ResNet is already heavily trained on object classification task.

Finally, table 7 shows the difference between the results in the work; show, attend and tell \cite{xu2015show} on flickr8k test-set against the best performing model in this work against the same test-set.

\begin{table}[htb]
\centering
\caption{Comparison against the state of the art English language}
\begin{tabular}{ccccc}
\textbf{Feature} & \textbf{B-2} & \multicolumn{1}{c}{\textbf{B-3}} & \multicolumn{1}{c}{\textbf{B-4}}  & \multicolumn{1}{c}{\textbf{METEOR}} \\ \hline
Show, attend and tell           &        \textbf{45.7}         &        \textbf{31.4}         &                            \textbf{21.3}         &   \textbf{20.3}                                                                      \\
This work           &       36.38          &     25.93            &                        15.09             &     14.23                                                                     \\ \hline                                    
\end{tabular}
\end{table}

\section{Conclusion}

The literature and experiments conducted in this work clearly show that the Arabic image captioning task is lacking behind when compared to English. Also, the absence of benchmarks and dataset slows down the research. Also, it can be concluded that MTL can be utilized to overcome shortage of data for a specific task and its potential is worth further exploration in this task. Also, the use of pre-trained word embeddings clearly enhanced the performance of image captioning which further proves their usefulness and the fact that NLP advances can contribute to the advances of image captioning.
\section*{Acknowledgments}

The authors would like to thank Mawdoo3.com company for granting permission to use their servers for conducting the experiments.

\bibliographystyle{unsrt}  
\bibliography{references}

\end{document}